\documentclass{article}

\usepackage{arxiv}

\usepackage[utf8]{inputenc} % allow utf-8 input
\usepackage[T1]{fontenc}    % use 8-bit T1 fonts
\usepackage{hyperref}       % hyperlinks
\usepackage{url}            % simple URL typesetting
\usepackage{booktabs}       % professional-quality tables
\usepackage{amsfonts}       % blackboard math symbols
\usepackage{nicefrac}       % compact symbols for 1/2, etc.
\usepackage{microtype}      % microtypography
\usepackage{lipsum}
\usepackage{graphicx}

\usepackage{amsmath, amsfonts, booktabs}

\graphicspath{ {./images/} }

\title{A Multi-scale Graph Signature for Persistence Diagrams based on Return Probabilities of Random Walks}

\author{
 Chau Pham\textsuperscript{\textasteriskcentered}\\
  Department of Computer Science\\
  Boston University\\
  Boston, USA \\
  \texttt{chaupham@bu.edu} \\
   \And
 Trung Dang\textsuperscript{\textasteriskcentered} \\
  Department of Computer Science\\
  Boston University\\
  Boston, USA \\
  \texttt{trungvd@bu.edu} \\
  \And
 Peter Chin \\
  Department of Computer Science\\
  Boston University\\
  Boston, USA \\
  \texttt{spchin@bu.edu} \\
}

\begin{document}

\newcommand{\theName}{RPNet}

\maketitle
\begingroup\renewcommand\thefootnote{\textasteriskcentered}
\footnotetext{Equal contribution
% ; This work was supported in part by NGA, grant number: HM047619C0011. PC would also like to gratefully acknowledge partial support for National Science Foundation.
}
\endgroup

\begin{abstract}
  Persistence diagrams (PDs), often characterized as sets of death and birth of homology class, have been known for providing a topological representation of a graph structure, which is often useful in machine learning tasks. Prior works rely on a single graph signature to construct PDs. In this paper, we explore the use of a family of multi-scale graph signatures to enhance the robustness of topological features. We propose a deep learning architecture to handle this set input. Experiments on benchmark graph classification datasets demonstrate that our proposed architecture outperforms other persistent homology-based methods and achieves competitive performance compared to state-of-the-art methods using graph neural networks. In addition, our approach can be easily applied to large size of input graphs as it does not suffer from limited scalability which can be an issue for graph kernel methods.
\end{abstract}

% keywords can be removed
%\keywords{First keyword \and Second keyword \and More}

\section{Introduction}
Persistent homology is a commonly used tool in the field of Topological Data Analysis, which is designed to track topological changes as data is examined at different scales. Its main descriptor is the persistence diagram (PD), often represented as a multiset of birth and death pairs of topological features. Since it can encode topological and geometrical properties of the data, it is potentially complementary to features retrieved by other classical descriptors. 
%PD is represented as a multiset of intervals in the Euclidean space, which can be utilized as features in machine learning tasks through constructing a kernel or vectorization. 
Applications of PD have been found in different fields, from signal analysis to 3D shape classification \cite{perea2015sliding,buchet2018persistent,li2014persistence,carriere2015stable,saggar2018towards,suh2019persistent,pranav2017topology, kannan2019persistent}. 

A recent line of research is primarily aimed at leveraging PD in the task of graph classification. Given a graph signature on vertices of the graph, one can construct a sublevel filtration from the nested sequence of increasing subgraphs. 
%Some commonly used graph signatures include node degree \cite{hofer2017deep}, Ricci-curvature \cite{zhao2019learning}, Jaccard-index\cite{zhao2019learning}, and heat kernel \cite{carriere2019perslay}. 
The PDs obtained from this filtration process can be used in machine learning tasks by constructing a kernel, such as sliced Wasserstein kernel \cite{carriere2017sliced} and Fisher kernel \cite{le2018persistence}, or associating them to a vector space, such as persistence landscape \cite{bubenik2015statistical} and persistence image \cite{adams2017persistence}. However, kernel methods require evaluation for each pair of graphs thus do not scale well to the size of the dataset, and vectorization methods are usually non-parametric or have a few trainable parameters, resulting in suffering from being agnostic to the type of application. For more scalability and flexibility, a deep learning model that operates on sets can be used directly on PDs \cite{hofer2017deep,carriere2019perslay}.

%Graphs are an important data modality which appear commonly in many real-world applications, such as social networks, chemical compounds, and biological structures. 
%One of the major tasks in graph data analysis is graph classification, which aims to determine the class a graph belongs to, such as classifying anatomical therapeutic chemicals~\cite{zhao2021convolutional}, identifying mutagenic compounds~\cite{li2021mutagenpred}. 

In this paper, we focus on graph classification using PDs from a deep learning perspective. A majority of prior works rely on a scalar node descriptor, such as node degree \cite{hofer2017deep} or heat kernel signature \cite{carriere2019perslay}, to construct the filtration, where the choice is often empirical. However, a multi-dimensional embedding is usually preferred to describe a graph node. While multi-dimensional persistence has been studied \cite{carlsson2009theory,lesnick2015theory}, it exhibits an essentially different character from its one-dimensional version and is often complicated to generalize corresponding concepts, such as persistence diagrams, making it nontrivial to utilize them in downstream models. 
Instead, we explore the potential of leveraging multiple PDs for each data sample based on a family of multi-scale graph signatures: the $k$-step return probabilities of random walks, where $k$ is the number of random steps from a starting node, indicating the size of the local structure being explored. For example, 2-hop random walk can help to capture the neighborhood size of surrounding nodes, while  3-hop can capture circles in the graph. The idea is that different scales help the model capture more information about its neighborhood, which in turn can help improve the model's performance. The return probability feature has been used as a node structural role descriptor to construct a kernel for graph classification \cite{zhang2018retgk}. We show that with an appropriate application and model architecture, it can also be useful in topology-based approaches.

%Prior works mostly focus on a single scalar descriptor function such as , geodesic or Euclidean distance~\cite{li2017metrics}, which may not be adequate to distill useful features from the graph. In this work, we utilize persistent homology to extract features of the graph across multiple scales to enhance the richness of the representation. The extraction can be used separately as an input or incorporated with other features of the graph for downstream tasks.  

%\begin{figure}[!htb]
%\centering
%\includegraphics[width=\linewidth]{img/flow_blackwhite.png}
%\caption{Overview of the persistence-based data analysis framework: From the graph input, we label its vertices by  $K$-dimensional vectors using $K$ return probabilities of random walks. We also introduce \theName{}, a neural network architecture to handle this type of persistence diagrams.}
%\label{fig:flow}
%\end{figure}

Our contributions are as follows. First, we propose a family of scalar graph signatures based on the return probability of random walks to construct multiple PDs per data sample. The PDs are padded to a fixed size and stacked together to form a multi-channel input. Second, we introduce \theName{}, a deep neural network architecture designed to learn from the proposed features. We demonstrate our method on a wide range of benchmark datasets and observe that it often outperforms other persistence diagram-based and neural network-based approaches, while being competitive with state-of-the-art graph kernel methods.
%\footnote{Our code will be released on Github}
% Our code is publicly available on Github\footnote{\url{https://github.com/chaudatascience/persistence_diagrams_gnn}}

%Our contributions are as follows. We first come up with a method to utilize persistent homology in the graph classification task. Instead of using a single graph signature to obtain persistence diagrams of a graph as in previous work, we consider a family of a multi-scale graph signature based on return probabilities of different random walks. From each of these probabilities, we generate a persistence diagram, which is then stacked together to form a multi-channel input.  This allows the machine learning model to access more useful information distilled from the graph. We also introduce \theName{}, a deep learning network architectures able to work efficiently on this type of dataset (Figure \ref{fig:flow}). Finally, we demonstrate our method on a wide range of benchmark datasets and observe that it achieves 1\% improvement in most of the datasets over previous persistence diagram-based and neural network based-approaches, and are comparably to state-of-the-art graph kernel methods.

The rest of this paper is organized as follows. Section \ref{sec:related_work} and Section \ref{sec:background} briefly go through some related work and background. Section \ref{sec:proposed_method} describes our proposed graph signature and network architecture. Section~\ref{sec:experiments} and Section~\ref{sec:results} present our experiments and results. Finally, we discuss future directions in Section~\ref{sec:conclusion}.

%--------------------------------------------------
%--------------------------------------------------
\section{Related Works}
\label{sec:related_work}

Since PDs have an unusual structure for machine learning approaches, many techniques have been proposed to map them into machine learning compatible representations. A popular technique is vectorization. For instance, Bubenik et al. \cite{bubenik2015statistical} propose mapping persistence barcodes into a Banach space, referred to as persistence landscape. Adams et al. \cite{adams2017persistence} propose persistence image, which is a discretization of the persistence surface. Another approach is constructing a kernel from PDs. For example, the sliced Wasserstein kernel \cite{carriere2017sliced}, or persistence Fisher kernel \cite{le2018persistence} are designed to work on PDs. Rieck et al. \cite{rieck2019persistent} define a kernel constructed from a vector representation of persistence diagram through the Weisfeiler-Lehman subtree feature. While these methods make it much easier to deploy machine learning techniques, they are pre-defined and thus often suboptimal to a specific task.

Learning representation has been investigated to provide a task-optimal representation in the vector space for PDs. Hofer et al. \cite{hofer2017deep} and Carriere et al. \cite{carriere2019perslay} propose a deep learning architecture designed to handle sets of points in PDs. Zhao et al. \cite{zhao2019learning} propose a weighted kernel for persistence image. These methods have been proven to perform well on a wide range of benchmarks.

In these works, a specific node descriptor is usually required for the filtration process. Some of the simple choices include node degree ~\cite{hofer2017deep} or the node attributes if available~\cite{rieck2019persistent}. More complicated descriptors include the heat kernel signature ~\cite{carriere2019perslay}, Ricci-curvature \& Jaccard-index ~\cite{zhao2019learning}. For non-attributed graphs, ~\cite{rieck2019persistent} uses a stabilization procedure based on node degree to leverage more topological information.

Our work employs a similar approach in \cite{hofer2017deep,carriere2019perslay} to propose a deep learning architecture that contains pooling layers to learn from multiset inputs. Different from prior works, our architecture is able to handle multiple PDs for each data sample, each computed from a node descriptor in the proposed family of graph signatures.

%Most of the recent works on persistent homology have mainly focused on scalar functions to calculate persistence diagrams. Ensembles of multiple signature functions (or multi-filtrations), based on the theory of multidimensional persistent homology~\cite{carlsson2009theory}~\cite{lesnick2015theory}, arise naturally and have been investigated in applications including shape and image classification~\cite{li2014persistence}~\cite{ adcock2014classification}. While using multiple persistence diagrams enhances the richness of invariants, it exhibits an essentially different character from its one-dimensional version and is often complicated to generalize corresponding representations, such as persistence barcodes or bottleneck distance.

\section{Background}
\label{sec:background}

We briefly mention the concepts 
% due to the space constraint 
and refer interested readers to~\cite{edelsbrunner2010computational, aktas2019persistence} for more detailed definitions of homology.

\subsection{Simplices and Simplicial Complexes}

\begin{figure}[t]
\centering
\includegraphics[width=0.5\linewidth]{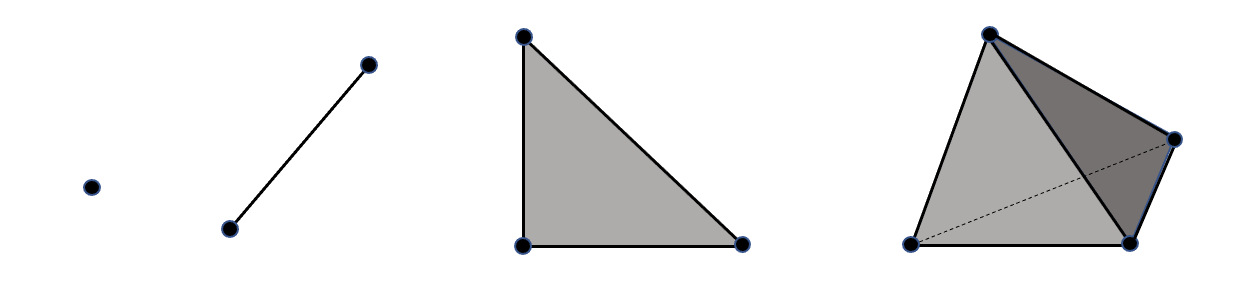}
\caption{Examples of simplices. From left to right: a vertex, an edge, a triangle, and a tetrahedron}
\label{fig:simplices}
\end{figure}

\paragraph*{Simplices} are higher dimensional analogs of points, line segments, and triangles, such as a tetrahedron. Formally, an $i$-simplex $\sigma$ is the convex hull  of $i+1$ affinely independent points, i.e., a set of all convex combinations $\lambda_{0} v_{0}+\lambda_{1} v_{1}+\ldots+\lambda_{i} v_{i}$ where $\lambda_{0}+\lambda_{1}+\ldots+\lambda_{i}=1$ and $\lambda_{j} \leq 1$ for all $j \in\{0,1, \ldots, i\} .$ For example, a \textit{0-simplex} is a single point $a_0$ (the vertex), a \textit{1-simplex} is a line segment $(a_0, a_1)$ (the edge), and a \textit{2-simplex} is a triangle $(a_0, a_1, a_2)$ with its interior (Figure \ref{fig:simplices}). Simplices are the building blocks of simplicial complexes.

\begin{figure}[t]
\centering
\includegraphics[width=0.7\linewidth]{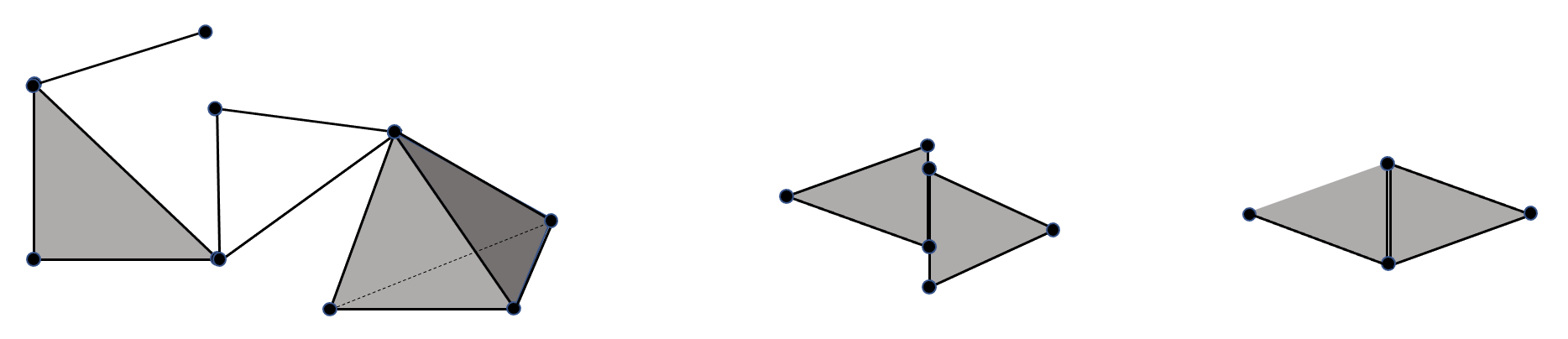}
\caption{left: simplicial complex created by 0-, 1-, and 2-simplicies; middle: not a simplicial complex since the intersection of the two triangles is not an edge; right: not a simplicial complex because of missing an edge}
\label{fig:simplicial_complex}
\end{figure}

\paragraph*{Simplicial Complexes} A simplicial complex is a finite collection of simplices $K$ such that (1) given a simplex $\sigma \in K$,  a face $\tau \leq \sigma$ implies $\tau \in K$, and (2) two simplices $\sigma_0, \sigma_{1} \in K$ implies $\sigma_0 \cap \sigma_{1}$ is either empty or a common face of both.
The dimension of $K$ is the maximum dimension of any of its simplices. The first condition states that if a simplex is in $K$, then its faces are also in $K$. For the second condition, we are only allowed to glue simplices by their common faces to avoid no improper intersections (Figure \ref{fig:simplicial_complex}).

\subsection{Persistence Diagrams of Graphs}

\begin{figure*}[!htb]
\centering
\includegraphics[width=\linewidth]{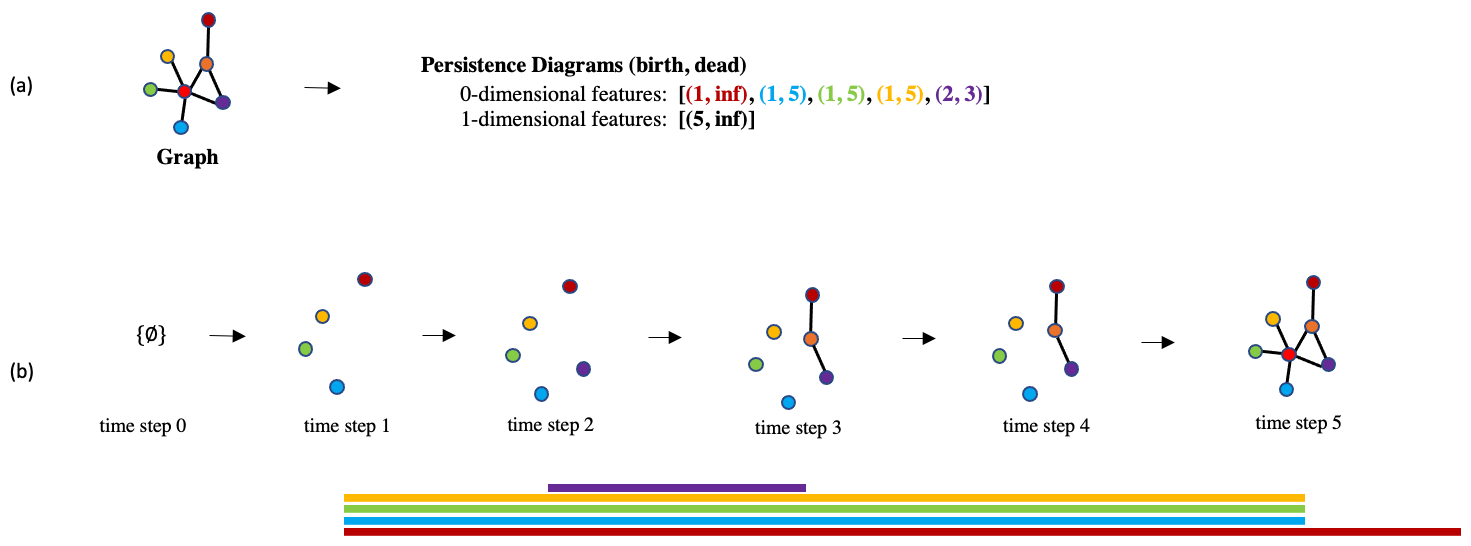}
\caption{0-dimensional and 1-dimensional persistent homology on a graph based on node-degree filtration. (a) Given an input graph, the persistence diagrams are computed by persistent homology;  (b) More details on the computation: The growth process of the graph by time steps and its corresponding bar codes. At the first time step (time step 0), there is no node and edge. At the next time steps, nodes and edges start to appear. The bar codes represent the persistence diagrams (birth, death). }
\label{fig:persistence_diagram}
\end{figure*}

Consider a graph $G=(V, E)$ with $n$ nodes, $m$ edges and a node descriptor function $\omega: V\rightarrow\mathbb{R}$. First, we use the descriptor to label the vertices in $G$. These labels are used as threshold values to define the sublevel graph $G_\alpha=(V_\alpha,E_\alpha)$ where $\alpha\in\mathbb{R}$, $V_\alpha=\{v\in V: \omega(v)\le\alpha\}$ and $E_\alpha=\{(u, v)\in E: u, v\in V_\alpha\}$. This provides a nested family of simplicial complexes $G_{\alpha_0}\subseteq G_{\alpha_1}\subseteq ...\subseteq G_{\alpha_{n}}\subseteq G_{\alpha_{n+1}}$ where $\alpha_0=-\infty$, $\alpha_{n+1}=+\infty$, $\alpha_1\le \alpha_2\le\cdots\le\alpha_n$ are weights assigned nodes sorted in increasing order. $(G_{\alpha_i})_{i=0}^{n+1}$, starting with the empty graph and ending with the full graph $G$, is called a filtration of $G$. During this construction, some holes may appear and disappear; in other words, the time of appearance and disappearance of topological features such as loops, and connected components are recorded. When a component first appears, its \textit{birth} is kept track of, and when it gets merged into another component, we record the \textit{death}. Such homological features can be considered the graph feature. Specifically, these intervals are persistence diagram that consists of the birth and death time of the feature as points (birth, death) in $\mathbb{R}^{2}$. Figure \ref{fig:persistence_diagram} illustrates  0-dimensional and 1-dimensional persistence barcodes and the corresponding persistence diagram of the filtration on a simple input graph. For each vertex $v_i$, its node degree is used as the node descriptor $\omega(v_i)=\sum_{j} \boldsymbol{A}_{i, j}$, where $\boldsymbol{A}$ is the adjacency matrix of $G$.

% \subsection{Graph signatures}

% \subsubsection{Laplacian-based}
% The symetric normalized graph Laplacian $L_{\omega}=L_{\omega}(G)$ is the linear operator acting on the space of functions defined on the vertices of $G$, and is represented by the matrix $L_{\omega}=I-D^{-\frac{1}{2}}AD^{\frac{-1}{2}}$. It admits an orthonormal basis of eigenfunctions $\Psi=\{\psi_1,...,\psi_n\}$ and its eigenvalues satisfy $0\le\lambda_1\le...\le\lambda_n\le 2$. The heat kernel signature with diffusion parameter $t$ is the function $\text{hks}_{G,t}$ defined on the vertices of G by $\text{hks}_{G,t}:v\mapsto\sum_{k=1}^n\exp(-t\lambda_k)\psi_k(v)^2$

% Random walk has been used  Let $p_{G,k}(\tau|i)$ denote the probability of a random walker on $v_k$ at time $\tau$ that starts from $v_i$: $p_G(\tau|i)=[p_{G,1}(\tau|i),...,p_{G,N}(\tau|i)]$. The shape of the point cloud $P_G(\tau)=\{p_G(\tau|1), p_G(\tau|2),...,p_G(\tau|N)\}$.

\subsection{Return Probabilities of Random Walks}

Consider an undirected graph $G=(V, E)$. Let $ \boldsymbol{A} $ be the adjacency matrix of $G$ and $ \boldsymbol{D}$ be the degree matrix, i.e. a diagonal matrix with $ \boldsymbol{D}_{i,i}=\sum_{j} \boldsymbol{A}_{i, j}$.

 A $k$-step walk starting from node $v_{i}$ is a sequence of nodes $\left\{v_{i}, v_{i+1}, v_{i+2}, \ldots, v_{i+k}\right\}$, with $\left(v_{j}, v_{j+1}\right) \in E$, where $i \leq j < i+k$. A random walk on $G$ is a Markov chain $\left(X_{0}, X_{1}, X_{2}, \ldots\right)$, whose transition probabilities are

$$
\operatorname{Pr}\left(X_{i+1}=v_{i+1} \mid X_{i}=v_{i}, \ldots, X_{0}=v_{0}\right)=\operatorname{Pr}\left(X_{i+1}=v_{i+1} \mid X_{i}=v_{i}\right)=\frac{\boldsymbol{A}_{i, j}}{\boldsymbol{D}_{i, i}}
$$

The transition probability matrix is calculated by $\boldsymbol{P}=\boldsymbol{D}^{-1} \boldsymbol{A}$. The $k$-hop transition probability matrix is $\boldsymbol{P}^k$, where $\boldsymbol{P}^k_{i,j}$ denotes the transition probability in $k$ steps from node $v_i$ to $v_j$. The return probability of random walks is defined as $P_{G,i}(k|i)$, which is the probability to return the source node after $k$ hops.  When $k=0$, all the probabilities are equal to $0$. 

A straightforward way to calculate a transition probability matrix $k$ requires $(k-1) \times n \times n$ matrix multiplication of $\boldsymbol{P}$, which is $O((k-1) n^{3})$ in time complexity. We are only interested in the diagonal elements, thus the matrix can be computed more efficiently and the time complexity can be reduced to $O\left(n^{3}+(k+1) n^{2}\right)$~\cite{zhang2018retgk}. The calculation of all the $k$-hop transition probability matrix with $k\in\{1,..., K\}$ takes time $O(n^3+(K+1)n^2)$, allowing computation to be scalable to large graphs.

\subsection{Deep Learning with Sets}
Most machine learning algorithms take fixed-size vectors as input. They are not designed for data in the form of sets. In contrast to vectors, sets are invariant to permutation and have a variable length. Thus, dealing with this type of input is a challenging and non-trivial task. There has been some research aimed at tackling this problem by employing permutation-invariant functions that have two desirable characteristics: invariant to the ordering of the inputs and ability to handle variable-size inputs~\cite{zaheer2017deep,rezatofighi2017deepsetnet,xu2018spidercnn}. Specifically, a permutation invariant function $L$ in $\mathbb{R}^{p}$ is defined 
 
 $$L\left(\left\{x_{1} \ldots x_{n}\right\}\right)=\rho\left(\sum_{i=1}^{n} \phi\left(x_{i}\right)\right)$$
 
 where $\phi: \mathbb{R}^{p} \rightarrow \mathbb{R}^{q}$, $n$ is the number of points, and $\rho: \mathbb{R}^{q} \rightarrow \mathbb{R}^{q}$. For example, average-pooling and sum-pooling are two pre-defined permutation invariant functions that are commonly used to handle sets~\cite{kipf2016semi, hamilton2017inductive, atwood2016diffusion}. 

% To tackle this problem, several neural network architectures have been proposed to work with them (e.g. \cite{zaheer2017deep,rezatofighi2017deepsetnet,xu2018spidercnn}). One common idea is to use permutation-invariant functions whose outputs do not depend on the order of the input, such as max-pooling to aggregate information from set elements into global features. Specifically, a permutation invariant function $L$ in $\mathbb{R}^{p}$ is satisfied 
% $L\left(\left\{x_{1} \ldots x_{n}\right\}\right)=\rho\left(\sum_{i=1}^{n} \phi\left(x_{i}\right)\right)$
% where $\phi: \mathbb{R}^{p} \rightarrow \mathbb{R}^{q}$, $n$ is the number of points, and $\rho: \mathbb{R}^{q} \rightarrow \mathbb{R}^{q}$.

%--------------------------------------------------
%--------------------------------------------------
\section{Proposed Method}
\label{sec:proposed_method}

In this section, we describe our approach to using the return probabilities of random walks for topological graph representation and propose our neural network architecture, referred to as \theName{}. Let $G=(V, E)$ be an undirected graph, where $V$ is the set of vertices and $E$ is the set of edges connecting two vertices, we define a set of $K$ descriptors $\text{RP}_k$ on $V$, where $\text{RP}_k(v)$ represents the $(k+1)$-hop return probabilities of random walk starting from the vertex $v$, with $k\in[1,K]$. The higher value of $k$, the larger a subgraph is considered. By definition:
\[\text{RP}_k(v_i)=\left[(\boldsymbol{D}^{-1}\boldsymbol{A})^k\right]_{i,i}\]
where $\boldsymbol{A}$ is the adjacency matrix, $\boldsymbol{D}$ is the degree matrix, and $v_i$ is the $i$-th vertex in $G$.

%\subsection{Feature Extraction from PDs}
%Given a graph as input, we label each of its vertices by using return probabilities of random walks. Then, the resulting persistence diagrams are stacked together to form a multi-channel input which is fed into a neural network model.

%\paragraph*{Vertex Labeling} We use $K$ return probabilities of random walks to provide a multi-resolution node descriptor for the graph. 
%We call this descriptor function RP$_K$: $V \rightarrow \mathbb{R}^{K}$ for all $(k+1)$-hop transitions, with $k\in\{1,..., K\}$,  $K\in\mathbb{N}$. Each of the vertices in the graph is labeled by RP$_K$ as a $K$-dimensional vector. The higher value of $K$, the larger a subgraph involving the source node is considered. 

\begin{figure}[!htb]
\centering
\includegraphics[width=0.6\linewidth]{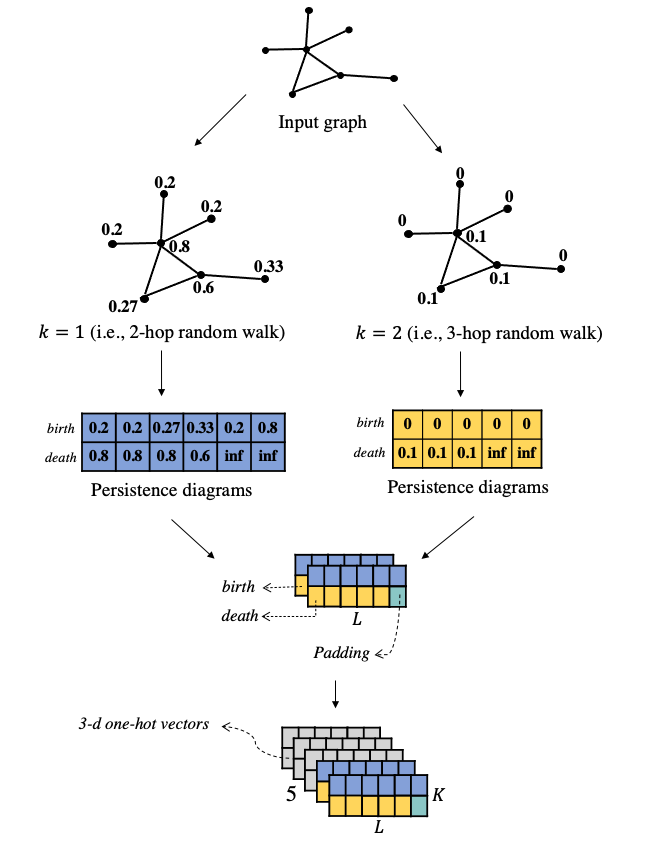}
\caption{Procedure to generate and process $K$ (e.g., $K=2$) persistence diagrams from a graph using return probabilities of random walks: We first use each return probability of random walk to label the nodes of the graph, which are used to generate pairs of (birth, death), then stack them together with 3-d one-hot vectors to obtain the features.}
\label{fig:vertex_labelling}
\end{figure}

\paragraph*{PDs Computation and Processing}Using the sublevel filtration based on each of $K$ node descriptors $\text{RP}_k$, we can compute $K$ 0,1-dimensional PDs for each graph $G$. We split points in these diagrams into three groups: \textit{0-homology essential points}  (0-homology points whose \textit{death} is infinite), \textit{0-homology non-essential points} (0-homology points whose \textit{death} is finite) and \textit{1-homology essential points} (1-homology points whose \textit{death} is infinite). Note that 1-homology does not disappear since we do not consider higher degree homology; therefore, \textit{1-homology non-essential point} (1-homology points whose death is finite) does not exist. We normalize these points to the range $[0, 1]$ by dividing them by the maximum coordinate (i.e., \textit{birth, death}) value ($+\infty$ is also mapped to 1), and add a 3-dimensional one-hot vector characterizing each group. Thus, each point is represented by a vector in $\mathbb{R}^5$: \textit{birth}, \textit{death}, and 3-d one-hot vector. After this process, we obtain $K$ sets of 5-dimensional vectors. These sets can vary in size (for example, in Fig. \ref{fig:vertex_labelling}, with $k=1$, the first set has 6 points (blue), while $k=2$ gives 5 points (yellow)), thus we pad them to have the same number of elements,  $L$, in each set. The final resulting output is a feature $\boldsymbol{X}\in\mathbb{R}^{K\times L\times 5}$ for each graph $G$. Figure \ref{fig:vertex_labelling} illustrates this process on a simple graph.

%\paragraph*{Multi-channel Input} For each graph and weight function defined by the $k$-hop return probability with $k=1,..., K$, we calculate the persistence diagram up to degree 1. Points in these diagrams are categorized into three groups: \textit{0-homology essential points}, \textit{0-homology non-essential points} and \textit{1-homology essential points} (note that 1-homology does not disappear; therefore, \textit{1-homology non-essential point} does not exist). Figure \ref{fig:vertex_labelling} illustrates the procedure to generate $K$ persistence diagrams from a graph. To represent them in the same space, we normalize all points to the range $[0, 1]$ ($+\infty$ is also mapped to 1) and add a 3-dimensional one-hot vector characterizing each group. Thus, each point is represented by a vector in $\mathbb{R}^5$. The output of this process is $K$ sets of 5-dimensional vectors.
%Consider a graph, we convert it into $K$-channel inputs, each denoted as $\textbf{X}_k\in\mathbb{R}^{L\times d}$,  where $k\in\{1,..., K\}$, $K$ is the number of persistence diagrams corresponding to $K$ return probabilities of random walks, $L$ is the maximum length of each diagram and $d=5$ is the feature dimension. We use padding to make all the diagrams have the same size $L$.

\subsection{Network architecture}

\begin{figure*}[!bth]
\centering
\includegraphics[width=\linewidth]{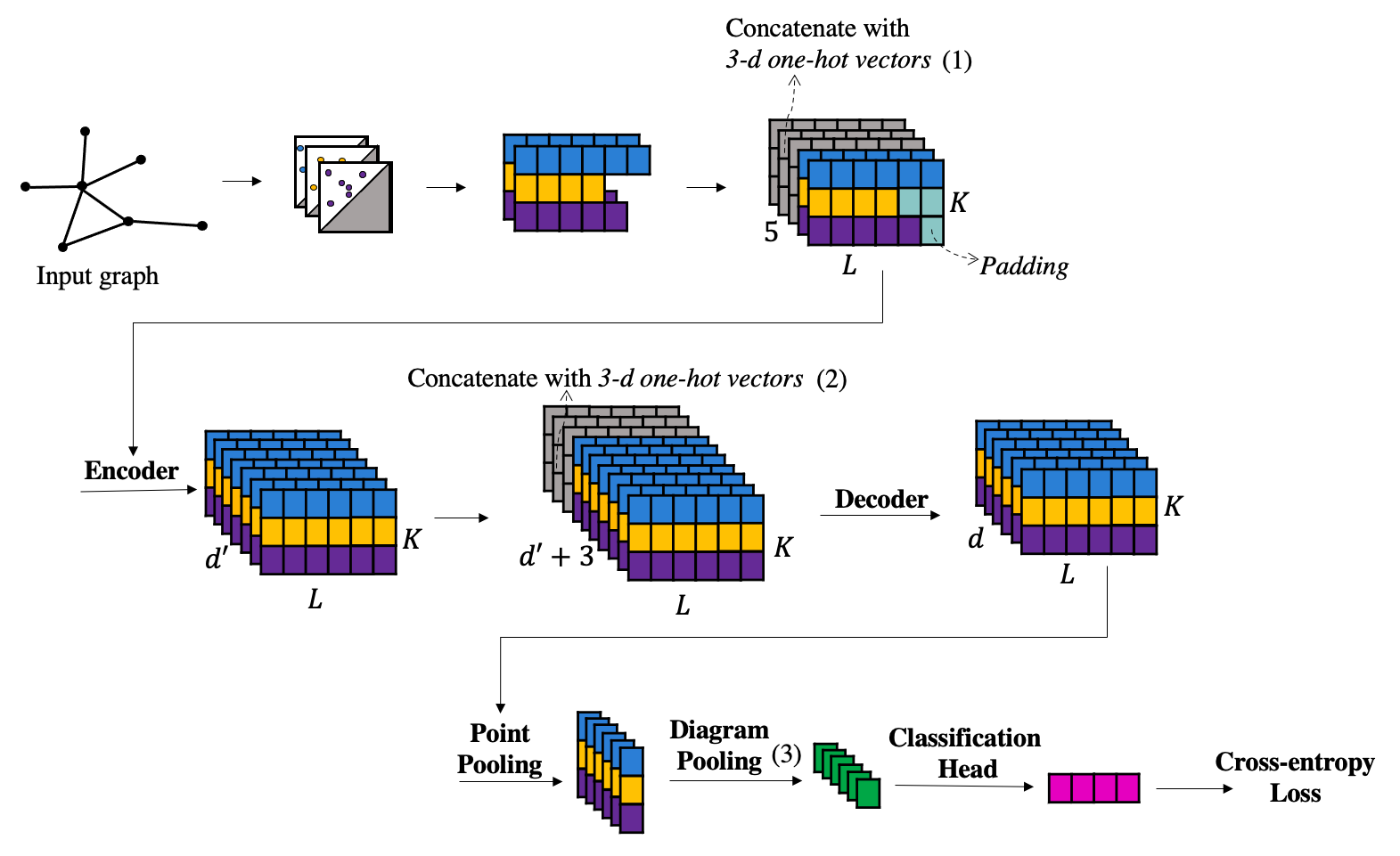}
\caption{Overview architecture of \theName{} - our proposed neural network}
\label{fig:RPNet}
\end{figure*}

We propose \theName{}, a neural network operating on the extracted features. The network consists of an encoder, a decoder, a point pooling layer, a diagram pooling layer, and a classification head. First, the encoder maps each point into a $d'$-dimensional space:

\[\boldsymbol{E}_{\text{point}}=\text{Encoder}(\boldsymbol{X})\in\mathbb{R}^{K\times L \times d'}\]

Next, we concatenate the encoded feature $\boldsymbol{E}_{\text{point}}$ with the 3-d one-hot vectors to obtain  $\boldsymbol{E'}_{\text{point}} \in\mathbb{R}^{K\times L\times (d'+3)}$. We then pass it to a decoder to map each point into $d$-dimensional space:
\[\boldsymbol{D}_{\text{point}}=\text{Decoder}(\boldsymbol{E'}_{\text{point}})\in\mathbb{R}^{K\times L \times d}\]

The point pooling layer performs sum pooling on points in each diagram, and the diagram pooling layer performs weighted pooling on diagrams for each graph.

% \[\boldsymbol{E}_{\text{diagram}}=\frac{1}{L}\sum_{l=1}^L \boldsymbol{E}_{\text{point}}[:, l, :]\]
\[\boldsymbol{X}_{\text{diagram}}=\sum_{l=1}^L \boldsymbol{D}_{\text{point}}[:, l, :]\]
\[\boldsymbol{X}_{\text{graph}}=\sum_{k=1}^K\text{softmax}(w)_k\boldsymbol{X}_{\text{diagram}}[k,:]\]

The classification head is a multi-layer perceptron on the embedding of each graph, followed by a softmax function to output the class probabilities.

\[\boldsymbol{Y}=\text{softmax}(\text{MLP}(\boldsymbol{X}_{\text{graph}}))\]

Our network is optimized with the cross-entropy loss. Figure \ref{fig:RPNet} demonstrates the architecture of our proposed neural network.

\section{Experiments}
\label{sec:experiments}

\paragraph{Setup}

We use persistence diagrams up to degree 1 for each $k=1, ..., K$ as inputs to our model. We follow the conventional settings for graph classification as described in previous works \cite{hofer2017deep,carriere2019perslay}.  Specifically, we perform a 10-fold cross-validation (9 folds for
training and 1 fold for testing)  for each run. We train each model for at most 500 epochs and report the average accuracy and standard deviation on the epochs that have the lowest loss of each fold. Training is terminated after 50 non-decreasing epoch losses. Adam optimizer~\cite{kingma2014adam} is used with an initial learning rate between the range of 0.001 to 0.01, and a decay factor of 0.5 after every 25 epochs up to 6 times. For the encoder and decoder, we use several blocks of four following components: Linear Layer, Norm Layer (e.g., BatchNorm~\cite{ioffe2015batch}, LayerNorm~\cite{ba2016layer}), Dropout~\cite{srivastava14a}, and Activation Layer (e.g., ReLu\cite{agarap2018deep}, ELU~\cite{clevert2015fast}) in that order. We fine-tune the number of blocks (from 2 to 5) and their dimensions in the encoder, along with the learning rate. We report the best scores for each variant of our models, denoted as \theName-$K$, where $K\in\{1,2,4,8\}$ represents the number of return probabilities of random walks that are used to generate the multi-scale graph representation. 
It is worth noting that all the baselines we compare with follow the same setup, thus it is a fair comparison.
% Dropout ~\cite{srivastava2014dropout} with dropout rate of 0.2 is used in the attention network.

\paragraph{Datasets}
\begin{table}[ht]
\caption{Descriptions of datasets used in our experiments}
\label{tab:dataset}
\vskip 0.15in
\begin{center}
%\begin{small}
\begin{sc}
\resizebox{.48\textwidth}{!}{\begin{tabular}{lcccc}
\toprule
                & \#graphs    & avg \#nodes         & avg \#edges         & \#classes\\
\midrule
MUTAG   &188 &17.93   &19.79   &2 \\
PROTEINS    &1113    &39.06   &72.82  & 2\\
NCI1    &4110       & 29.87      & 32.30 &  2\\
NCI109  &  4127    &     29.68   & 32.13  & 2\\
PTC\_FM   & 349     &   14.11    &   14.48  & 2\\
PTC\_FR   & 351    & 14.56      &15.00   &2\\
PTC\_MM    &   336  &    13.97  &     14.32&   2\\
PTC\_MR    &344      &  14.29  &  14.69   &2\\
COX2      &467      & 41.22  &  43.45   &2\\
DHFR   & 756     &42.43   & 44.54   &2\\

REDDIT-BINARY   & 2000          & 429.6                 & 497.8                 & 2         \\
IMDB-BINARY     & 1000           & 19.77	                & 96.53                 & 2         \\
COLLAB          & 5000          & 74.49                 & 2457.78	            & 3         \\
IMDB-MULTI      & 1500          & 13.00 	            &65.94                  & 3         \\
REDDIT-5K       & 4999          & 508.5                 & 594.9                 & 5         \\
REDDIT-12K      & 11929         & 391.4                 & 456.9                 & 12        \\

\bottomrule
\end{tabular}}

\end{sc}
%\end{small}
\end{center}
\vskip -0.1in
\end{table}
We evaluated our proposed method in graph classification tasks on 16 graph benchmark datasets from different domains such as small chemical compounds or protein molecules (e.g., COX2, PROTEIN), and social networks (e.g., REDDIT-BINARY, IMDB-M). We group these datasets into two categories: (1) Datasets with less than 1,000 graphs, or less than 100 vertices per graph on average, and (2) Datasets with more than 1,000 graphs and more than 100 vertices per graph on average.
% We evaluated our proposed method in graph classification task on 16 graph benchmark datasets from different domains such as small chemical compounds or protein molecules (e.g., COX2, PROTEIN), and social networks (e.g., REDDIT-BINARY, IMDB-M). 
It is worth noting that while most of the datasets consist of an adjacency matrix, some bioinformatic datasets also provide categorical node attributes. The aggregate statistics of these datasets are provided in Table \ref{tab:dataset}\footnote{More details can be found at  \url{https://ls11-www.cs.tu-dortmund.de/staff/morris/graphkerneldatasets}}.

% \bgroup
%\def\arraystretch{0.5}
% \setlength\tabcolsep{2pt}

\begin{table*}[ht!]

\caption{Classification results (accuracy and standard deviation) on dataset group 1. The best-performing results are highlighted in boldface. Note  $^1$: graph kernel, $^2$: graph neural network, $^3$: method based on persistence homology. We use "-" to indicate that the dataset was not used in the baseline paper. }
\label{tab:result1}
\vskip 0.15in
\begin{center}

\begin{small}
\begin{sc}
\resizebox{\textwidth}{!}{\begin{tabular}{lcccccccccc}

\toprule
                & MUTAG          & PROTEINS          & NCI1          & NCI109            & PTC-MR            & PTC-FR            & PTC-MM            & PTC-FM            & COX2          & DHFR                  \\
\midrule
WL$^1$          & 82.1          & -             & 82.2          & 82.5              & -                 & -                 & -                 & -                 & -             & -                     \\
RetGK$^1$       & 90.3$\pm$1.1     & 75.8$\pm$0.6     & 84.5$\pm$0.2     & -                 & 62.5$\pm$1.6         & 66.7$\pm$1.4         & 65.6$\pm$1.1         & 62.3$\pm$1.0         & -             & -                     \\
GCAPS$^2$       & -             & \textbf{76.4$\pm$4.1}     & 82.7$\pm$2.4     & 81.1$\pm$1.3         & -                 & -                 & -                 & -                 & -             & -                     \\
PersLay$^{2,3}$ & 89.8$\pm$0.9          & 74.8$\pm$0.3          & 73.5$\pm$0.3          & 69.5$\pm$0.3              & -                 & -                 & -                 & -                 & 80.9$\pm$1.0          & 80.3$\pm$0.8                  \\
SV$^{1,3}$      & 88.2$\pm$1.0     & 72.6$\pm$0.4     & 71.3$\pm$0.4     & 69.8$\pm$0.2         & -                 & -                 & -                 & -                 & 78.4$\pm$0.4     & 78.8$\pm$0.7             \\
P-WL$^{1,3}$    & 86.1$\pm$1.4     & 75.3$\pm$0.7     & 85.3$\pm$0.1     & 84.8$\pm$0.2         & 63.1$\pm$1.5         & 67.3$\pm$1.5         & 68.4$\pm$1.2         & 64.5$\pm$1.8         & -             & -                     \\
P-WLC$^{1,3}$   & 90.5$\pm$1.3     & 75.3$\pm$0.4     & \textbf{85.5$\pm$0.2}     &\textbf{ 85.0$\pm$0.3}         & \textbf{64.0$\pm$0.8}         & 67.2$\pm$1.1         & \textbf{68.6$\pm$1.8}         & \textbf{65.8$\pm$1.2}         & -             & -                     \\
\midrule
RPNet-1      & 89.4$\pm$4.0         & 74.5$\pm$4.5     & 73.5$\pm$1.1     & 70.7$\pm$1.6         & 60.8$\pm$5.3        & 66.7$\pm$3.6              & 64.9$\pm$5.5              & 63.3$\pm$6.8               & 81.8$\pm$3.0 & 76.3$\pm$3.6\\
RPNet-2      & 90.9$\pm$3.3 &75.8$\pm$3.3 &73.9$\pm$2.5 &71.1$\pm$1.7  & 60.7$\pm$4.6       & \textbf{68.1$\pm$7.3}              & 66.7$\pm$4.8     &63.3$\pm$5.3         & \textbf{82.4$\pm$2.7} & 75.5$\pm$4.0\\
RPNet-4      & \textbf{91.0$\pm$3.4}         & 75.4$\pm$2.6     & 72.1$\pm$1.8     & 70.0$\pm$1.9       & 62.5$\pm$6.1& 66.7$\pm$4.0        &63.7$\pm$4.9              & 64.5$\pm$6.1            & 81.6$\pm$3.5   & \textbf{81.3$\pm$3.0}\\
RPNet-8    & 91.0$\pm$4.0 &74.2$\pm$2.8      & 72.6$\pm$1.9     & 70.4$\pm$1.6        & 59.6$\pm$7.1          & 66.4$\pm$4.6 & 65.5$\pm$4.5              & 62.2$\pm$7.8             & 79.9$\pm$2.7  & 80.7$\pm$2.6\\
\bottomrule

\end{tabular}
}

\end{sc}
\end{small}

\end{center}
\vskip -0.1in
\end{table*}
\begin{table*}[!htbp]
\caption{Classification results  (accuracy and standard deviation) on dataset group 2. The best performances of our methods and other methods are highlighted in boldface. Note: "-" indicates the dataset was not used in the baseline paper.}
\label{tab:result2}
\vskip 0.15in
\begin{center}
\begin{small}
\begin{sc}
\begin{tabular}{llccccccccccc}
\toprule
               & & COLLAB        & RD-B      & RD-M5K    & RD-M12K   & IMDB-B        & IMDB-M\\
\midrule
Neural Net & GCAPS       & 77.7$\pm$2.5  & 87.6$\pm$2.5  & 50.1$\pm$1.7  & -             & 71.7$\pm$3.4  & 48.5$\pm$4.1  \\
&PersLay         & 76.4$\pm$0.4          & -             & 55.6$\pm$0.3          & 47.7$\pm$0.2          & 71.2$\pm$0.7          & 48.8$\pm$0.6          \\
&DeepTopo        & -             & -             & 54.5          & 44.5          &               &               \\

\midrule
Ours &RPNet-1            & 69.2$\pm$2.9 & 91.4$\pm$1.8 & 56.9$\pm$2.2 & 48.5$\pm$0.8 & 69.6$\pm$4.2 & 45.9$\pm$3.9 \\
&RPNet-2            & 70.0$\pm$2.2 & 91.6$\pm$2.0 & \textbf{57.6$\pm$2.4} &\textbf{48.5$\pm$1.0} & 68.3$\pm$3.9 & 46.8$\pm$3.1 \\
&RPNet-4            & 71.8$\pm$2.5 & \textbf{91.7$\pm$1.9} & 57.2$\pm$2.1 & 48.4$\pm1.0$ & \textbf{71.9$\pm$ 4.4} & 47.7$\pm$2.9\\
&RPNet-8            & \textbf{73.4$\pm$1.8} & 90.4$\pm$2.1 & 56.4$\pm$2.1 & 48.3$\pm$1.2 &  70.7$\pm$3.8 & \textbf{48.9$\pm$2.9} \\

\midrule
Graph Kernel & RetGK           & \textbf{81.0$\pm$0.3}  & \textbf{92.6$\pm$0.3}  & 56.1$\pm$0.5  &\textbf{ 48.7$\pm$0.2}  & 71.9$\pm$1.0  & 47.7$\pm$0.2  \\
& WKPI & -             & -             & \textbf{59.5$\pm$0.6}          & 48.4$\pm$0.5          &  \textbf{75.1$\pm$1.1}             & 49.5$\pm$0.4               \\
& SV              & 79.6$\pm$0.3  & 87.8$\pm$0.3  & 53.1$\pm$0.2  & -             & 74.2$\pm$0.9  &\textbf{ 49.9$\pm$0.3} \\

% RP              & -             & 91.7$\pm$1.2  & 57.0$\pm$3.0  & 49.2$\pm$1.5  & 65.9\pm3.4 & 46.5\pm4.8 \\
% RP-W            & -             & 92.8$\pm$1.1  & 57.3$\pm$2.6  & 48.8$\pm$1.7  & -             & -             \\
% RP-A            & -             & -             & -             & -             & -             & -             \\
\bottomrule
\end{tabular}
\end{sc}
\end{small}
\end{center}
\vskip -0.1in
\end{table*}

% \paragraph{Preprocessing}
% Return probabilities up to $K=8$ are calculated for every node in the graphs. Points in each persistence diagram are scaled across all samples.
% For essential points where the dead values are  $+\infty$, we map them to 1.

\paragraph{Baselines}
We compare our results with WL subtree~\cite{shervashidze2011weisfeiler}, RetGK~\cite{zhang2018retgk}; deep learning-based approaches  GCAPS~\cite{verma2018graph}, PersLay~\cite{carriere2019perslay}, DeepTopo~\cite{hofer2017deep}; Kernel-based methods with topological features: SV~\cite{tran2018scale}, P-WL and P-WLC~\cite{rieck2019persistent}, WKPI~\cite{zhao2019learning}. 
For the above-mentioned baselines, we report their accuracies and standard deviations (if available) in the original papers, following the same evaluation setting. Some of our datasets were not used in certain papers (but used by others), we denote this by "-" or leave out the method if it did not use any of the datasets in the resulting table. 

\paragraph{Implementation}
Models are implemented in Pytorch~\cite{paszke2019pytorch}. The calculation of persistence diagrams are performed by Dionysus~\footnote{\url{https://mrzv.org/software/dionysus2}} and Gudhi~\footnote{\url{http://gudhi.gforge.inria.fr}} libraries.

\section{Results}
\label{sec:results}

\paragraph*{Dataset Group 1} Table \ref{tab:result1} demonstrates the results on 10 graph datasets. Our method performs the best on 4 out of 10 datasets. On the other 6 datasets, it performs comparably with PersLay \cite{carriere2019perslay}, another persistence diagram based method. 

\paragraph*{Dataset Group 2} For large datasets, our method is ranked in the top 3 on most of the datasets as observed in Table \ref{tab:result2}. When compared against other methods in Neural Network family, it outperforms these methods on 5 out of 6 datasets.

Some baseline methods such as RetGK~\cite{zhang2018retgk} leverage node attributes (node labels) of the graph for classification purposes. Our results are promising considering that the approach does not use any node attributes as the features. When comparing the best model among our variants (i.e., different values of $K$), we observe that using more than 1 return probability of random walk achieves better results in most of the datasets. In particular, RPNet-1 performs the best only on 1 dataset, in a total of 16 datasets. This implies that our method benefits from using multiple PDs based on multi-scale graph signatures. On the other hand, while our variant with $K = 2$ (RPNet-2) works the best on small datasets in group 1, we observe that increasing $K$ (i.e., $K = 4$ and $K = 8$) gains some improvement on large datasets in group 2. Despite having high performance on most benchmarks, graph kernel methods require computing and storing evaluations for each pair of graphs in the dataset, making them intractable when dealing with datasets containing a large number of graphs. For example, the random walk kernel (RW)~\cite{vishwanathan2010graph} takes more than 3 days on \textit{NCI1} dataset \cite{zhang2018end}. Our approach can avoid the quadratic complexity with respect to the number of graphs required for graph kernels, and thus can easily apply to real-world graph datasets. It is also worth mentioning that our approach has higher variance compared with others, such as kernel methods. This high variance is a common problem when using deep learning approaches, which can be observed in deep learning baselines such as GCAPS~\cite{verma2018graph}.

\paragraph*{Ablation experiments} In this section, we conduct some ablation study on our proposal's architecture. We start with our RPNet-4, and then remove some of its components to study their efficiency. The comparison is shown in Table \ref{tab:ablation}.
 
We can observe that removing 3-d one-hot vectors (1) and (2) decreases the accuracy by  6.4\%, 6.6\%, and 2.3\% on \textit{NCI1, IMDB-M, PROTEINS} respectively. The main efficiency mainly comes from one-hot vectors (1) as it contributes much more to the performance compared with one-hot vectors (2). This shows the importance of using one-hot vectors to keep the information about the type of persistence diagrams from the original graphs. Besides, we find that learning a set of weights to aggregate the diagram features is beneficial to the model. When compared with the average pooling as a baseline, we find that weighted diagram pooling improves approximately 1-3\% accuracy on the 3 datasets.
 
According to the experimental results, incorporating all the three components into our model, in general, can also help to reduce the variance, allowing us to obtain more stable performance.
\begin{table}[ht]
\caption{Ablation experiment results of our model's architecture on 3 datasets \textit{NCI1, IMDB-M, PROTEINS}.
Note: \textit{one-hot (1) } denotes concatenation of the 3-d one-hot vectors with topological features extracted from the graph,
\textit{one-hot (2) } denotes concatenation of the 3-d one-hot vectors with encoded features (after encoding step), and \textit{weighted diagram pooling (3)} indicates using weighted sum to aggregate the diagrams, otherwise average pooling is employed. These components are indicated by (1), (2), and (3) in Fig. \ref{fig:RPNet}, respectively.}
\label{tab:ablation}
\vskip 0.15in
\begin{center}
%\begin{small}
\begin{sc}
\resizebox{.8\textwidth}{!}{\begin{tabular}{lcccc}
\toprule
Dataset & one-hot (1)       & one-hot (2)      & weighted diagram pooling (3) & Accuracy $\pm$ std  \\ 
\hline
NCI1  &                  &                & $\checkmark$             & 65.7$\pm$2.9       \\
       &                  &  $\checkmark$  & $\checkmark$             & 66.6 $\pm$2.6     \\
       & $\checkmark$     &                & $\checkmark$             & 70.8$\pm$1.5       \\
       & $\checkmark$     & $\checkmark$   &                          & 70.7$\pm$1.0       \\
       & $\checkmark$     & $\checkmark$   & $\checkmark$             & \textbf{72.1$\pm$1.8}       \\

\hline
IMDB-M  &                   &              & $\checkmark$             & 41.1$\pm$4.1       \\
        &                   & $\checkmark$ & $\checkmark$             & 41.2$\pm$4.1     \\
        & $\checkmark$      &              & $\checkmark$             & 43.4$\pm$3.2       \\
        & $\checkmark$      & $\checkmark$ &                          & 44.3$\pm$3.7       \\
        & $\checkmark$      & $\checkmark$ & $\checkmark$             & \textbf{47.7$\pm$2.9}        \\

\hline
PROTEINS &                   &              & $\checkmark$             & 73.1$\pm$3.6       \\
        &                   & $\checkmark$ & $\checkmark$             & 73.1$\pm$2.8     \\
        & $\checkmark$      &              & $\checkmark$             & 73.1$\pm$3.8       \\
        & $\checkmark$      & $\checkmark$ &                          & 72.3$\pm$4.2       \\
        & $\checkmark$      & $\checkmark$ & $\checkmark$             & \textbf{75.4$\pm$2.6}        \\

\bottomrule
\end{tabular}}
\end{sc}
\end{center}
\vskip -0.1in
\end{table}

% \section{Discussion}
% \label{sec:discussion}

% For \theName{}, we observe that persistence diagrams of a large graph are usually duplicated. Thus, we find out that processing these diagrams with a frequency coefficient speeds up the training while producing the same results. We apply the trick to facilitate the training process. In terms of the architecture, we also try to add a residual connection and then increase the depth of networks. Besides, we conduct some experiments on replacing the weighted pooling with an attention-based~\cite{vaswani2017attention} pooling layer, which is an extension of the weighted pooling layer where weights are obtained from a pair-wise scoring function between elements in the set. While weighted poolings learn and share a fixed weight for each diagram, attention poolings calculate this weight every time based on the pair-wise interaction between persistence diagrams. However, these techniques do not show a clear improvement as it turns out that the simpler version of \theName{} works better.

\section{Conclusion}
\label{sec:conclusion}
In this paper, we propose a family of multi-scale graph signatures for persistence diagrams based on return probabilities of random walks. We introduce RPNet, a deep neural network architecture to handle the proposed features and demonstrate its efficiency on a wide range of benchmark graph datasets. We show that our proposed method outperforms other persistent homology-based methods and graph neural networks while being on par with state-of-the-art graph kernels. There are some other interesting directions worth exploring. One of those is to employ a kernel-based approach with our multiset features. There has been some work on defining the distance between two persistence diagrams. It would be promising to build a kernel method working on the proposed multiscale representation. Besides, our approach does not make use of the node attributes, which may be critical to classifying graphs, such as chemical compounds where each node attribute represents a chemical element. Thus, another future direction is to investigate in detail how to incorporate the node attribute into our multi-scale graph signature. We are also interested in developing a theory for multi-dimensional persistent homology and using it for deep learning models in downstream tasks.

%%
%% The acknowledgments section is defined using the "acks" environment
%% (and NOT an unnumbered section). This ensures the proper
%% identification of the section in the article metadata, and the
%% consistent spelling of the heading.

% TODO: edit this
\section{Acknowledgments}
Approved for public release, 22-607 

%%
%% The next two lines define the bibliography style to be used, and
%% the bibliography file.
\bibliographystyle{unsrt} 
\bibliography{my_bib}

\end{document}